\PassOptionsToPackage{unicode}{hyperref}
\PassOptionsToPackage{hyphens}{url}
\documentclass[
  12pt,
]{article}
\usepackage{amsmath,amssymb}
\usepackage{lmodern}
\usepackage{iftex}
\ifPDFTeX
  \usepackage[T1]{fontenc}
  \usepackage[utf8]{inputenc}
  \usepackage{textcomp} 
\else 
  \usepackage{unicode-math}
  \defaultfontfeatures{Scale=MatchLowercase}
  \defaultfontfeatures[\rmfamily]{Ligatures=TeX,Scale=1}
\fi
\IfFileExists{upquote.sty}{\usepackage{upquote}}{}
\IfFileExists{microtype.sty}{
  \usepackage[]{microtype}
  \UseMicrotypeSet[protrusion]{basicmath} 
}{}
\usepackage{xcolor}
\usepackage[margin=1in]{geometry}
\usepackage{graphicx}
\makeatletter
\def\maxwidth{\ifdim\Gin@nat@width>\linewidth\linewidth\else\Gin@nat@width\fi}
\def\maxheight{\ifdim\Gin@nat@height>\textheight\textheight\else\Gin@nat@height\fi}
\makeatother
\setkeys{Gin}{width=\maxwidth,height=\maxheight,keepaspectratio}
\makeatletter
\def\fps@figure{htbp}
\makeatother
\newlength{\cslhangindent}
\newlength{\csllabelwidth}
\newlength{\cslentryspacingunit} 
\newenvironment{CSLReferences}[2] 
 {
  \setlength{\parindent}{0pt}
  \ifodd #1
  \let\oldpar\par
  \def\par{\hangindent=\cslhangindent\oldpar}
  \fi
  \setlength{\parskip}{#2\cslentryspacingunit}
 }%
 {}
\usepackage{calc}

\usepackage{accents}
\usepackage{amsmath}
\usepackage{bm}
\usepackage{booktabs}
\usepackage{braket}
\usepackage{caption}
\usepackage{epigraph}
\usepackage{extarrows}
\usepackage{float}
\usepackage[bottom,flushmargin,hang,multiple]{footmisc}
\usepackage{graphicx}
\usepackage{hyperref}
\usepackage[export]{adjustbox}
\usepackage{libertine}
\usepackage[libertine]{newtxmath}
\usepackage[mathcal]{euscript}
\usepackage{mathrsfs}
\usepackage{multirow}
\usepackage{scrextend}
\usepackage{setspace}
\usepackage{soul}
\usepackage{subcaption}
\usepackage{titlesec}
\usepackage[utf8]{inputenc}
\ifLuaTeX
  \usepackage{selnolig}  
\fi
\IfFileExists{bookmark.sty}{\usepackage{bookmark}}{\usepackage{hyperref}}
\IfFileExists{xurl.sty}{\usepackage{xurl}}{} 
\hypersetup{
  pdftitle={The Creative Frontier of Generative AI: Managing the Novelty-Usefulness Tradeoff},
  hidelinks,
  pdfcreator={LaTeX via pandoc}}

\title{The Creative Frontier of Generative AI: Managing the
Novelty-Usefulness Tradeoff}
\author{}
\date{\vspace{-2.5em}}

\begin{document}
\maketitle

\begin{center}
\author
{Anirban Mukherjee,$^{1}$ Hannah H. Chang$^{2\ast}$\\
\medskip
\normalsize{$^{1}$Samuel Curtis Johnson Graduate School of Management, Cornell University,}\\
\normalsize{Sage Hall, Ithaca, NY 14850, USA}\\
\normalsize{$^{2}$Lee Kong Chian School of Business, Singapore Management University,}\\
\normalsize{50 Stamford Road, Singapore, 178899}\\
\normalsize{$^\ast$To whom correspondence should be addressed; E-mail: hannahchang@smu.edu.sg.}\\
}
\medskip
\date{}
\end{center}
\medskip

\noindent 

\noindent In this paper, drawing inspiration from the human creativity
literature, we explore the optimal balance between novelty and
usefulness in generative Artificial Intelligence (AI) systems. We posit
that overemphasizing either aspect can lead to limitations such as
hallucinations and memorization. Hallucinations, characterized by AI
responses containing random inaccuracies or falsehoods, emerge when
models prioritize novelty over usefulness. Memorization, where AI models
reproduce content from their training data, results from an excessive
focus on usefulness, potentially limiting creativity. To address these
challenges, we propose a framework that includes domain-specific
analysis, data and transfer learning, user preferences and
customization, custom evaluation metrics, and collaboration mechanisms.
Our approach aims to generate content that is both novel and useful
within specific domains, while considering the unique requirements of
various contexts.

\begin{center}\rule{0.5\linewidth}{0.5pt}\end{center}


\newpage

\doublespacing

Creativity, a defining trait of human intelligence, is a subject of
extensive study and ongoing discussion
(\protect\hyperlink{ref-amabile1996creativity}{Amabile 1996},
\protect\hyperlink{ref-boden2004creative}{Boden et al. 2004}). Its
manifestations, such as divergent thinking that generates novel ideas
and convergent thinking that refines these ideas to meet specific goals,
have fueled numerous theories about its essence and underlying
processes. As artificial intelligence (AI) technologies continue to
advance, there is increasing interest in mirroring or simulating this
sophisticated, uniquely human form of creativity within generative AI
models. But why is this interest in AI creativity so significant? Is it
merely about making AI more human-like, or does embedding creativity
have pragmatic benefits for society?

The importance of creativity in AI goes beyond making machines mimic
human-like traits. AI models endowed with creative capabilities can
unlock a plethora of practical applications, from creating engaging
content in the entertainment industry to devising innovative solutions
in science, medicine, and business. Creativity in AI can generate
unforeseen solutions to complex problems, opening up new opportunities
and pathways that might not have been visible or conceivable through
human thought alone.

Indeed, early neural networks took cues from the intricate workings of
biological systems. Similarly, more advanced AI and machine learning
systems can gain from drawing parallels with social systems. These
systems, typified by dynamic interactions, collaboration, and
innovation, provide a rich source of inspiration for pushing AI
technologies forward. Human creativity, in particular---with its nuanced
interplay of novelty and usefulness---provides a valuable blueprint for
developing AI models capable of both innovating and delivering practical
solutions (\protect\hyperlink{ref-simonton2010creativity}{Simonton
2010}).

We bring to focus the vital elements of human creativity---novelty and
usefulness---in this paper. Novelty refers to generating original
content, previously unencountered, while usefulness pertains to crafting
content that is relevant, valuable, and practically applicable in a
given context. Striking this balance in AI models is a challenging
endeavor. Unlike traditional rule-based systems, AI models learn from
examples they encounter during their training phase. They do not have
access to an explicitly codified set of rules that separate fact from
fiction, leading to the generation of creative content that may either
deviate too far from practical constraints, known as hallucination, or
stay too rigidly within the confines of existing data, a phenomenon
known as memorization.

To tackle this challenge, we propose an approach inspired by human
creative processes. By incorporating principles like divergent and
convergent thinking, domain-specific creativity, evaluation, and
explanation, we aim to promote the development of AI models that
generate innovative content, minimize inaccuracies, and address ethical
considerations.

In the sections that follow, we will explore the conceptual definition
of creativity, delve into the challenge of balancing novelty and
usefulness in AI, and outline our proposed framework for the development
of generative AI models that are creative, useful, and ethical. By
drawing connections between human creativity and AI, we aim to stimulate
further discussion and research in the broader scientific community on
AI creativity.

\hypertarget{creativity-a-balance-of-novelty-and-usefulness}{%
\section{Creativity: A Balance of Novelty and
Usefulness}\label{creativity-a-balance-of-novelty-and-usefulness}}

Creativity is characterized as the capacity to generate novel, yet
valuable ideas or solutions to complex, open-ended tasks
(\protect\hyperlink{ref-amabile1983social}{Amabile 1983}). It requires
producing outcomes that are original, and at the same time, relevant,
feasible, or aligned with a specific goal. Creative tasks can be
heuristic, algorithmic, or a mixture of both. Heuristic tasks lack a
direct solution and call for innovative approaches, whereas algorithmic
tasks have a clear path to a single correct solution. For instance,
designing a fresh advertising campaign for a product is a heuristic task
since there is no one correct answer or predefined path to success;
instead, a range of creative solutions could potentially be effective.
On the other hand, solving a mathematical equation is algorithmic due to
its pre-defined resolution path.

\hypertarget{balancing-novelty-and-usefulness-in-real-world-examples}{%
\subsection{Balancing Novelty and Usefulness in Real-World
Examples}\label{balancing-novelty-and-usefulness-in-real-world-examples}}

\hypertarget{example-1-airbnbs-disruption-of-the-hospitality-industry}{%
\paragraph{Example 1: Airbnb's Disruption of the Hospitality
Industry}\label{example-1-airbnbs-disruption-of-the-hospitality-industry}}

Within Amabile's two-component model of creativity, Airbnb stands as an
exceptional example. The novelty of Airbnb's model upended traditional
hospitality norms. Before Airbnb, the concept of individuals renting out
their homes to strangers was largely foreign. This radical idea not only
diverged from the norm, but it also created an entirely new sector
within the hospitality industry. As for usefulness, Airbnb addressed
real-world demands by providing an income source for property owners and
a wide-ranging, affordable lodging choice for travelers. Notably, Airbnb
fostered a sense of community, typically absent in traditional
hospitality establishments. Thus, Airbnb's idea is not just novel; it
also efficiently solves a problem, demonstrating an ideal blend of
novelty and usefulness in the creative process.

\hypertarget{example-2-crispr-cas9s-gene-editing-revolution}{%
\paragraph{Example 2: CRISPR-Cas9's Gene Editing
Revolution}\label{example-2-crispr-cas9s-gene-editing-revolution}}

CRISPR-Cas9's introduction illustrates creativity within the
biotechnology sphere, aligning well with Amabile's model. The novelty of
CRISPR is clear: it repurposed a bacterial defense mechanism into a tool
for precise genetic editing, a groundbreaking approach. But novelty
alone doesn't justify CRISPR's broad success and influence; its
usefulness is just as vital. CRISPR-Cas9 has been used in the treatment
of genetic diseases such as sickle cell disease and beta thalassemia. It
also has potential applications in enhancing crop yield and resistance
and furthering genetic research. In simple terms, CRISPR-Cas9 not only
introduces a novel method, but it also effectively tackles critical
challenges across diverse fields.

\hypertarget{example-3-googles-pagerank-algorithm}{%
\paragraph{Example 3: Google's PageRank
Algorithm}\label{example-3-googles-pagerank-algorithm}}

Google's introduction of the PageRank algorithm brought a completely
novel approach to web search. Traditional search engines ranked results
based on the frequency of a search term appearing on a webpage. Google's
founders, Larry Page and Sergey Brin, proposed a new method, ranking
webpages based on their interconnectedness or link structure,
essentially viewing a webpage as important if other important pages
linked to it.

The usefulness of this novel approach became apparent as it resulted in
more accurate search results, making it easier for users to find
pertinent information. Google's novel approach to search transformed the
way people interacted with the internet, and it cemented Google's
position as a global tech giant. The introduction of Google's PageRank
algorithm brought a wholly novel approach to web search, significantly
enhancing its usefulness.

\hypertarget{generative-ai-navigating-the-novelty-usefulness-spectrum}{%
\section{Generative AI: Navigating the Novelty-Usefulness
Spectrum}\label{generative-ai-navigating-the-novelty-usefulness-spectrum}}

Generative artificial intelligence (AI) poses a challenge: balancing
novelty and usefulness, two crucial factors determining the success and
applicability of produced content. To navigate this spectrum more
effectively, AI models must carefully apply divergent and convergent
thinking and address hurdles associated with implicit learning and the
ambiguous information boundary. Here, we'll explore these core aspects
of generative AI and discuss potential strategies for mitigating ethical
concerns and biases.

\hypertarget{implicit-learning-and-the-ambiguous-information-boundary}{%
\subsubsection{Implicit Learning and the Ambiguous Information
Boundary}\label{implicit-learning-and-the-ambiguous-information-boundary}}

A significant challenge that arises in achieving the optimal balance
between novelty and usefulness in generative AI is due to the inherent
tension between these two aspects of creativity. Unlike traditional
rule-based systems, which explicitly encode information, generative AI
models engage in implicit learning. This means they analyze vast amounts
of data points in their training datasets to capture patterns and
relationships, thereby understanding domain constraints and principles
(\protect\hyperlink{ref-radford2019language}{Radford et al. 2019}).

Implicit learning could be likened to learning a language through
immersion, rather than formal education. An AI model, like a person
living in a foreign country, picks up on linguistic patterns, rules, and
idiosyncrasies through exposure to a wealth of data points
(conversations, written text, etc.), rather than being taught the rules
of grammar explicitly.

However, this form of learning also creates an ambiguous information
boundary, where it's challenging to distinguish between established
facts and potentially new information. This could be compared to our
foreign language learner beginning to invent words or phrases that sound
plausible but aren't actually part of the language. This blurred
boundary complicates the balance between novelty and usefulness, as the
model navigates a vast space of possibilities without explicit
guidelines on accuracy, truth, or appropriateness.

For instance, consider a language-based AI model trained on a vast
dataset from various internet sources, including scientific articles,
novels, news, and forums. When asked to generate an article about a
futuristic concept such as ``underwater cities,'' the AI model might end
up generating a highly creative and elaborate narrative, filled with
details about the city's architecture, transportation, and lifeforms.

However, because these details have been `imagined' by the model (which
essentially means the details are extrapolations from its training data
and not based on real, factual information), some of the content might
be scientifically inaccurate or purely speculative. For instance, the AI
model might suggest that these underwater cities are built from a
non-existent, corrosion-resistant material or inhabited by fictional
marine species. This is an example of hallucination, where the AI's
quest for novelty overrides the factual constraints of reality.

Similarly, suppose we have a generative AI model tasked with creating a
unique story. However, instead of generating a novel narrative, the
model produces a passage that is remarkably similar to a famous opening
of a well-known novel. For instance, it might generate a story beginning
with ``It was the best of times, it was the worst of times,'' closely
mirroring the opening line of Charles Dickens's ``A Tale of Two
Cities.''

This instance demonstrates memorization, where the AI model, in its
pursuit of creating useful (or contextually relevant) content, ends up
generating outputs that lack originality and essentially echo fragments
from its training data. In such cases, the AI model's creativity is
stifled, and the balance between novelty and usefulness is skewed
towards the latter.

\hypertarget{managing-novelty-and-usefulness-in-generative-ai}{%
\subsubsection{Managing Novelty and Usefulness in Generative
AI}\label{managing-novelty-and-usefulness-in-generative-ai}}

Balancing novelty and usefulness poses a significant challenge for
generative AI models. If these systems veer too far towards novelty,
they may override domain facts, principles, or boundaries, thereby
creating content that appears creative but significantly deviates from
the domain's foundational logic or history. This deviation could lead to
the compromise of its usefulness.

A phenomenon termed hallucination comes into play here. It involves AI
responses that contain random inaccuracies or falsehoods expressed with
unjustifiable confidence. A notable example of this is large-scale
language models like ChatGPT, which might unintentionally generate
outputs that blur the lines between reality and imagination, thus
stretching the boundaries of innovation but risking the generation of
misleading or fabricated information
(\protect\hyperlink{ref-brown2020language}{Brown et al. 2020}).

Hallucinations can be viewed as an emergent property of the creative
process. They represent not a failure of the model, but a misdirection
of emphasis. When users require factual information, the concept of
appropriateness takes precedence, and the model should adhere to the
truth reflected in the training examples rather than creating its own
reality. Hallucinations crop up when the model fails to achieve the
right balance.

Conversely, when AI systems lean heavily towards usefulness, they may
become overly fixated on generating content that strictly adheres to
real-world constraints and principles. This focus can lead to a
phenomenon known as memorization, where AI models reproduce content
verbatim from their training data. As a result, the generated outputs
may be useful but lacking in originality, potentially constraining their
creative potential.

\hypertarget{addressing-ethical-concerns-and-biases-in-generative-ai}{%
\subsubsection{Addressing Ethical Concerns and Biases in Generative
AI}\label{addressing-ethical-concerns-and-biases-in-generative-ai}}

The balancing act between novelty and usefulness is crucial for
addressing ethical concerns and biases in generative AI. It's not always
about limiting the generation of novel content; rather, it's ensuring
that AI-generated outputs are clearly labeled as novel, avoiding
confusion with established facts.

This balance can be achieved by introducing a novelty index for
AI-generated outputs. This index would enable users to differentiate
between novel content and known facts, reducing the potential negative
consequences of misleading content like deepfakes. The novelty index
could be calculated by comparing the generated outputs with the training
data. Outputs that closely mirror a training sample could be marked as
less novel, while outputs that significantly differ from a training
sample could be marked as more novel.

For instance, if an AI model generates a futuristic concept for a
zero-emissions vehicle powered by an entirely new form of energy, the
novelty index might rate this idea highly, indicating its divergence
from the established facts present in the model's training data. On the
other hand, a detailed explanation of an existing electric vehicle model
would likely receive a low novelty score, signaling its close alignment
with established knowledge.

Biases present in the training data could lead to skewed definitions of
appropriateness, creating outputs that reflect and amplify societal
biases (\protect\hyperlink{ref-crawford2017trouble}{Crawford 2017}). To
counter these biases, some researchers have moved away from allowing
biased data to define usefulness. Possible solutions include explicitly
labeling training data or adjusting the data distribution to emphasize
underrepresented groups (\protect\hyperlink{ref-sun2019mitigating}{Sun
et al. 2019}). Alternatively, implementing fairness-aware algorithms
such as adversarial debiasing can be effective
(\protect\hyperlink{ref-zhang2018mitigating}{Zhang et al. 2018}).
Adversarial debiasing, for example, trains an AI model to predict an
outcome while minimizing its ability to predict a protected attribute
(e.g., race or gender) from the model's predictions. This approach
lessens the impact of biases in the training data, ensuring the
generation of more balanced and fair outputs.

\hypertarget{gleaning-insights-from-human-creativity}{%
\section{Gleaning Insights From Human
Creativity}\label{gleaning-insights-from-human-creativity}}

\hypertarget{the-role-of-divergent-and-convergent-thinking}{%
\paragraph{The Role of Divergent and Convergent
Thinking}\label{the-role-of-divergent-and-convergent-thinking}}

Fundamental to the process of creativity is the fine equilibrium struck
between divergent and convergent thinking
(\protect\hyperlink{ref-boden2004creative}{Boden et al. 2004}).
Divergent thinking sets the stage for the generation of an array of
ideas or possible solutions, while convergent thinking shines in the
curation and selection of the best-fit idea or solution. Ample research
on creativity underpins the relative efficacy of these thinking styles,
establishing their pertinence to the context and stage of
problem-solving---divergent thinking gains prominence during ideation,
whereas convergent thinking finds utility in the evaluation and
selection phase.

Analogously, the pursuit of equilibrium between novelty and usefulness
in generative AI can be guided by principles of creative
problem-solving. This involves the identification of problems, the birth
of alternative solutions via divergent thinking, and the selection of
the most promising solution using convergent thinking. Imbuing
generative AI models with these creative problem-solving techniques can
engender structured and target-driven methodologies for content
generation.

Collaboration, a cornerstone of creativity literature, can bolster both
types of thinking in AI development. Techniques such as ensemble
learning or multi-agent systems can instill a collaborative spirit,
further optimizing the balance between novelty and usefulness in
AI-generated content. By fostering cooperation among AI models, a broad
spectrum of ideas can be generated, with another specialized set of
models focusing on their refinement in terms of usefulness.

Cognitive flexibility, defined as the capacity to alternate between
different modes of thinking, is underscored in creativity literature as
an indispensable factor in creative problem-solving
(\protect\hyperlink{ref-davis2009understanding}{Davis 2009}). In the AI
sphere, this might translate to the development of models capable of
dynamically modulating their emphasis between novelty and usefulness,
contingent upon task requirements or user preferences. Such flexible AI
models could deliver a more tailored and effective balance between
divergent and convergent thinking, ultimately enhancing the quality of
content.

A pragmatic approach to integrate divergent and convergent thinking into
generative AI models is a two-step process. Initially, models emulate
divergent thinking by promoting diversity and generating a broad set of
potential outputs. Subsequently, through convergent thinking, models
apply constraints or evaluation metrics to refine and select the most
fitting output. By architecturally incorporating these steps into
generative AI models, developers can facilitate the desired balance
between the generation of novel outputs and the usefulness and relevance
of the final content.

\hypertarget{the-significance-of-domain-specific-creativity}{%
\paragraph{The Significance of Domain-Specific
Creativity}\label{the-significance-of-domain-specific-creativity}}

Optimally balancing novelty and usefulness in generative AI models
necessitates adaptation to unique contexts or domains. Given the varying
degrees of novelty and usefulness across domains, AI models can be
custom-fit to generate content that is both imaginative and pertinent.
The creativity literature provides rich insights into this trade-off,
crucial for tailoring AI models.

One strategy involves incorporating domain-specific datasets during the
training phase or employing transfer learning techniques (i.e., adapting
pre-trained AI models for specific tasks)
(\protect\hyperlink{ref-pan2010survey}{Pan and Yang 2010}). This equips
AI models with a robust understanding of domain-specific language,
norms, and constraints, enabling them to cater to the diverse
requirements of various domains, such as healthcare or art.

A nuanced understanding of the specific needs for novelty and usefulness
within a context is pivotal for fine-tuning AI models. This can be
achieved by analyzing creativity literature to glean domain-specific
heuristics or evaluation criteria that reflect the ideal balance between
novelty and usefulness. These criteria, when integrated into the AI
model's objective function or fine-tuning process, enable
general-purpose AI models like ChatGPT to adapt more effectively. This
approach mirrors the development of neural networks inspired by
biological systems, anchoring AI models in human creativity systems.

Enhancing AI models' adaptability across contexts can be further
achieved by allowing users to specify their output preferences. This can
be realized through user interfaces that enable users to tweak
parameters related to novelty and usefulness, or by integrating user
feedback during the fine-tuning process of the AI model. By giving users
a voice to express their preferences, the balance between novelty and
usefulness can be tuned to their needs, generating AI content that meets
the specific requirements of different domains.

\hypertarget{evaluation-and-explanation}{%
\paragraph{Evaluation and
Explanation}\label{evaluation-and-explanation}}

The choice of evaluation metrics to assess generative AI models
critically influences the balance between novelty and usefulness.
Prioritizing grammatical correctness and fluency metrics may lead AI
models to produce coherent, yet unimaginative outputs. Conversely,
emphasizing uniqueness might lead to novel but grammatically incorrect
or difficult-to-understand outputs.

Researchers can cultivate balanced generative AI models by learning from
creativity literature and adopting evaluation techniques that consider
both novelty and usefulness. One such technique is the Consensual
Assessment Technique (CAT), which involves expert judgments to evaluate
the creativity of AI-generated outputs
(\protect\hyperlink{ref-amabile1982social}{Amabile 1982}). CAT offers a
comprehensive evaluation by amalgamating various aspects, including
fluency, coherence, and originality, into the assessment process
(\protect\hyperlink{ref-eastwood2018framework}{Eastwood and Williams
2018}).

Moreover, researchers can devise custom evaluation metrics that ensure a
balance between novelty and usefulness by amalgamating existing metrics.
For instance, they could blend metrics like BLEU, which measures the
similarity between the generated text and a reference text, with metrics
such as Self-BLEU, which evaluates the diversity of generated outputs by
comparing them to each other
(\protect\hyperlink{ref-zhang2018mitigating}{Zhang et al. 2018}). Such a
combination nudges AI models to generate content that strikes a balance
between coherence and originality. AI explainability and
interpretability, crucial for comprehending AI models' underlying
mechanisms and decision-making processes, can provide further support
for researchers in adjusting models and comprehending the
context-specific trade-offs between novelty and usefulness
(\protect\hyperlink{ref-gilpin2018explaining}{Gilpin et al. 2018}).

\hypertarget{a-framework-for-optimal-creativity-in-generative-ai}{%
\section{A Framework for Optimal Creativity in Generative
AI}\label{a-framework-for-optimal-creativity-in-generative-ai}}

We propose a multifaceted framework to construct AI systems that strike
an optimal balance between novelty and usefulness. This approach
amalgamates domain-specific knowledge, user preferences, and cooperative
techniques. Our proposed model includes the following pillars:

\begin{enumerate}
\def\labelenumi{\arabic{enumi}.}
\item
  \textbf{Domain-Specific Analysis}: To tailor AI models to cater to the
  unique requirements of different contexts, a comprehensive
  understanding of the domain-specific characteristics and constraints
  is required. Through a deep dive into the relevant creativity
  literature and consultation with domain experts, we can establish the
  desired equilibrium between novelty and usefulness. This ensures the
  AI-generated content resonates with the specifics of the domain while
  maintaining its innovative edge.
\item
  \textbf{Domain-Specific Data and Transfer Learning}: The application
  of domain-specific datasets during the training phase, coupled with
  transfer learning methodologies, can fine-tune AI models to suit the
  target domain. This nuanced understanding of the domain-specific
  language, norms, and constraints results in more relevant and novel
  content generation.
\item
  \textbf{User Preferences and Customization}: Introducing a user
  interface or feedback mechanism that allows users to express their
  novelty-usefulness preferences or adjust AI model parameters enhances
  the personalization of the experience. This facilitates the adaptation
  of AI-generated content to meet the varying needs and specifications
  of different domains.
\item
  \textbf{Custom Evaluation Metrics}: The development of bespoke
  evaluation metrics that encapsulate both novelty and usefulness can
  enhance the effectiveness of AI models. These metrics, inspired by
  existing ones and human judgement techniques such as the Consensual
  Assessment Technique, allow the models to generate content that
  harmonizes originality and applicability.
\item
  \textbf{Collaboration Mechanisms}: The implementation of collaborative
  mechanisms, such as ensemble learning or multi-agent systems, refines
  the balance between novelty and usefulness in AI-generated content. By
  encouraging diverse idea generation and refinement based on utility
  through a collective of AI models, the end content is more likely to
  fulfill the dual objectives of creativity.
\end{enumerate}

\hypertarget{conclusion}{%
\section{Conclusion}\label{conclusion}}

In this paper, we delve into the challenges faced by generative AI
models in striking an optimal balance between novelty and usefulness.
Inspired by human creativity, we propose a comprehensive approach that
leverages creative problem-solving methods to enhance these models'
capabilities.

The balance between novelty and usefulness isn't just crucial for AI
creativity---it's also key to addressing ethical concerns and biases. By
carefully managing this tradeoff, we can guide AI models to produce
creative outputs that not only respect real-world constraints and
principles but are also self-aware and appropriately labeled for
novelty. Additionally, confronting biases in the training data is
critical for promoting fairness.

Navigating this balance is vital to aligning generative AI models with
user expectations and ensuring they make positive contributions to
society. As the field of AI continues to evolve, we hope our research
serves as a valuable link between human and AI creativity.

\hypertarget{references}{%
\section{References}\label{references}}

\hypertarget{refs}{}
\begin{CSLReferences}{1}{0}
\leavevmode\vadjust pre{\hypertarget{ref-amabile1982social}{}}%
Amabile TM (1982) Social psychology of creativity: A consensual
assessment technique. \emph{Journal of personality and social
psychology} 43(5):997.

\leavevmode\vadjust pre{\hypertarget{ref-amabile1983social}{}}%
Amabile TM (1983) The social psychology of creativity: A componential
conceptualization. \emph{Journal of personality and social psychology}
45(2):357.

\leavevmode\vadjust pre{\hypertarget{ref-amabile1996creativity}{}}%
Amabile TM (1996) \emph{Creativity and innovation in organizations}
(Harvard Business School Boston).

\leavevmode\vadjust pre{\hypertarget{ref-boden2004creative}{}}%
Boden MA et al. (2004) \emph{The creative mind: Myths and mechanisms}
(Psychology Press).

\leavevmode\vadjust pre{\hypertarget{ref-brown2020language}{}}%
Brown T, Mann B, Ryder N, Subbiah M, Kaplan JD, Dhariwal P, Neelakantan
A, et al. (2020) Language models are few-shot learners. \emph{Advances
in neural information processing systems} 33:1877--1901.

\leavevmode\vadjust pre{\hypertarget{ref-crawford2017trouble}{}}%
Crawford K (2017) The trouble with bias. \emph{Conference on neural
information processing systems, invited speaker}.

\leavevmode\vadjust pre{\hypertarget{ref-davis2009understanding}{}}%
Davis MA (2009) Understanding the relationship between mood and
creativity: A meta-analysis. \emph{Organizational behavior and human
decision processes} 108(1):25--38.

\leavevmode\vadjust pre{\hypertarget{ref-eastwood2018framework}{}}%
Eastwood C, Williams CK (2018) A framework for the quantitative
evaluation of disentangled representations. \emph{International
conference on learning representations}.

\leavevmode\vadjust pre{\hypertarget{ref-gilpin2018explaining}{}}%
Gilpin LH, Bau D, Yuan BZ, Bajwa A, Specter M, Kagal L (2018) Explaining
explanations: An overview of interpretability of machine learning.
\emph{2018 IEEE 5th international conference on data science and
advanced analytics (DSAA)}. (IEEE), 80--89.

\leavevmode\vadjust pre{\hypertarget{ref-pan2010survey}{}}%
Pan SJ, Yang Q (2010) A survey on transfer learning. \emph{IEEE
Transactions on knowledge and data engineering} 22(10):1345--1359.

\leavevmode\vadjust pre{\hypertarget{ref-radford2019language}{}}%
Radford A, Wu J, Child R, Luan D, Amodei D, Sutskever I, et al. (2019)
Language models are unsupervised multitask learners. \emph{OpenAI blog}
1(8):9.

\leavevmode\vadjust pre{\hypertarget{ref-simonton2010creativity}{}}%
Simonton DK (2010) Creativity in highly eminent individuals. \emph{The
Cambridge handbook of creativity}:174--188.

\leavevmode\vadjust pre{\hypertarget{ref-sun2019mitigating}{}}%
Sun T, Gaut A, Tang S, Huang Y, ElSherief M, Zhao J, Mirza D, Belding E,
Chang KW, Wang WY (2019) Mitigating gender bias in natural language
processing: Literature review. \emph{arXiv preprint arXiv:1906.08976}.

\leavevmode\vadjust pre{\hypertarget{ref-zhang2018mitigating}{}}%
Zhang BH, Lemoine B, Mitchell M (2018) Mitigating unwanted biases with
adversarial learning. \emph{Proceedings of the 2018 AAAI/ACM conference
on AI, ethics, and society}. 335--340.

\end{CSLReferences}

\hypertarget{acknowledgements}{%
\section{Acknowledgements}\label{acknowledgements}}

This research was supported by the Ministry of Education (MOE),
Singapore, under its Academic Research Fund (AcRF) Tier 2 Grant,
No.~MOE-T2EP40221-0008.

\end{document}